**Sketch-Based Facade Renovation With Generative AI Models**

*A Streamlined Framework for Bypassing As-Built Modelling in Industrial Adaptive Reuse*

Warissara Booranamaitree[1#], Xusheng Du[2#], Yushu Cai[3], Zhengyang Wang[4], Ye Zhang[5*], and Haoran Xie[6,7*]
[1]*Chulalongkorn University.*
[2,4,6]*Japan Advanced Institute of Science and Technology.*
[3,5]*Tianjin University.*
[7]*Waseda University.*
[1]*6532152721@student.chula.ac.th, 0009-0003-0588-3260*
[2]*s2320034@jaist.ac.jp ([#]equal contribution), 0009-0008-1086-2081*

**Abstract.** Facade renovation offers a more sustainable alternative to full demolition, yet producing design proposals that preserve existing structures while expressing new intent remains challenging. Current workflows typically require detailed as-built modelling before design, which is time-consuming, labour-intensive, and often involves repeated revisions. To solve this issue, we propose a three-stage framework combining generative artificial intelligence (AI) and vision-language models (VLM) that directly processes rough structural sketch and textual descriptions to produce consistent renovation proposals. First, the input sketch is used by a fine-tuned VLM model to predict bounding boxes specifying where modifications are needed and which components should be added. Next, a stable diffusion model generates detailed sketches of new elements, which are merged with the original outline through a generative inpainting pipeline. Finally, ControlNet is employed to refine the result into a photorealistic image. Experiments on datasets and real industrial buildings indicate that the proposed framework can generate renovation proposals that preserve the original structure while improving facade detail quality. This approach effectively bypasses the need for detailed as-built modelling, enabling architects to rapidly explore design alternatives, iterate on early-stage concepts, and communicate renovation intentions with greater clarity.

**Keywords.** Industrial building renovation, vision-language model, diffusion model, user sketches, facade renovation

# 1. Introduction

A leading proponent of industrial heritage revitalization, Norman Foster advocated that such buildings should be given new life through well-planned design interventions (Kim et al., 2018). Compared with demolition and reconstruction, adaptive reuse and facade renovation offer more sustainable and economically viable approaches to transforming industrial facilities. However, traditional industrial facade renovation follows a linear and labour-intensive process, progressing from conceptual sketches to 3D models and construction drawings. Each step requires manual intervention and repeated revisions, resulting in high labour and communication costs (Sun et al., 2022).

In recent years, generative artificial intelligence (AI) has emerged as a promising tool for architectural renovation, offering efficient and creative solutions for updating existing structures (Li et al., 2025). Despite its advantages, applying generative AI directly to industrial renovation remains challenging for several reasons. First, existing models struggle to understand architectural semantics and cannot automatically identify which regions of a facade should be modified, added, or preserved, making it difficult to translate design intent into spatial action. Second, although diffusion-based image generation can produce visual elements, it often fails to maintain structural coherence, stylistic consistency, and fine-grained photorealistic details, resulting in fragmented or unrealistic outcomes that limit professional design communication.

To address these issues, we propose a three-stage generative AI framework that integrates vision–language models (VLMs) for semantic understanding of rough structural sketches and textual descriptions, with image generation models, Stable Diffusion, IP-Adapter, and ControlNet, to produce contextually coherent and structurally consistent renovation proposals (Figure 1). We focus on the adaptive transformation of industrial factories into commercial buildings, which retains industrial aesthetics while improving facade visual quality. Experiments and case studies show that our framework generates renovation proposals that maintain structural integrity, enrich facade articulation, and achieve photorealistic realism, offering a practical solution for industrial building renewal.

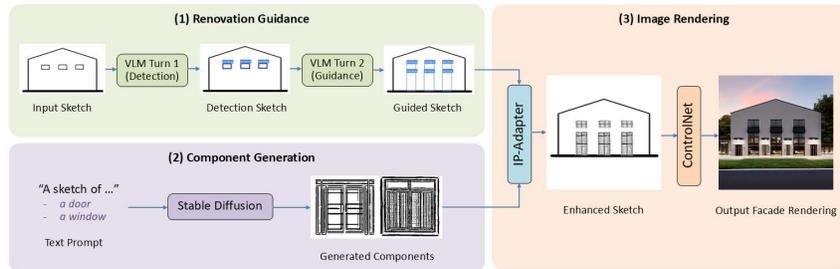

*Figure 1. Overview of the Three-Stage Framework*

The main contributions of this work are summarized as follows:

- We propose a three-stage generative framework that integrates sketches and textual descriptions to automate facade renovation design.

- We construct a publicly accessible dataset of industrial buildings sketches that includes renovated building sketches paired with corresponding original sketches.

- We demonstrate that incorporating a VLM model with architectural sketches can combine both textual and visual reasoning, resulting in meaningful design guidance.

## 2. Related Work

### 2.1. ARCHITECTURAL DESIGN AND FACADE RENOVATION

Recent advances in generative AI have introduced diverse models for architectural image generation and visual exploration. Text-to-image diffusion models such as Stable Diffusion (Rombach et al., 2022) enable high-quality synthesis from natural language prompts, while conditional variants like ControlNet (Zhang et al., 2023) incorporate structural priors such as edge or depth maps to enhance controllability. Image-to-image translation frameworks such as Pix2Pix (Isola et al., 2017) have been applied to tasks like facade editing and style transfer, and adapter-based techniques such as IP-Adapter (Ye et al., 2023) further improve control by introducing external visual or stylistic references. However, these methods remain largely disconnected from architectural practice in that they focus on aesthetic generation rather than existing built conditions, structural context, or early-stage facade workflows. In contrast, research on building renovation within architecture fields has primarily concentrated on structural adaptation, sustainability, and energy performance rather than automation or semantic design generation. Studies on the adaptive reuse of heritage buildings emphasise preserving historical character while improving functionality and environmental performance (Li et al., 2021), yet these processes remain highly dependent on manual design input, site-specific expertise, and deterministic evaluation frameworks. This gap highlights the need for data-driven generative methodologies that can automate early-stage design exploration. By integrating sketch-conditioned inputs with generative and adapter-based models, the proposed framework enables high-fidelity renovation visualisations that realistically render new facade elements while preserving the geometric and semantic integrity of existing buildings.

### 2.2. VISION-LANGUAGE MODELS

A vision–language model is a multimodal AI framework that integrates a vision encoder with a large language model (LLM) to jointly process and reason across visual and textual modalities. Early works such as CLIP (Radford et al., 2021) aligned image and text embeddings within a shared latent space, enabling powerful zero-shot recognition and cross-modal understanding. Building on this foundation, recent large-scale systems such as Qwen3-VL (Bai et al., 2025) further enhance multimodal reasoning by coupling pretrained visual encoders with language models capable of fine-grained spatial grounding and semantic interpretation across diverse tasks including visual question answering, image captioning, and image–text retrieval. In this study, a fine-tuned Qwen3-VL is employed to interpret rough architectural sketches and textual prompts, serving as a semantic bridge that guides the generative process toward structurally consistent and semantically aligned facade renovation designs.

## 3. Method

As shown in Figure 1, the proposed framework comprises three sequential stages that collectively enable effective facade renovation from input sketches. First, a fine-tuned VLM analyses the input sketch and provides renovation guidance. Then, a diffusion model generates new structural components according to the VLM's guidance and integrates them into the sketch. Finally, the enhanced sketch is refined into a photorealistic facade image through a ControlNet-based rendering process.

## 3.1. RENOVATION GUIDANCE VIA VISION-LANGUAGE MODEL

The proposed generation framework takes a rough structural sketch that preserves the overall geometry and spatial proportion of a building as input, and aims to produce textual renovation guidance describing where modifications should occur and what structural components should be added, such as new windows, entrances, and facade features. To achieve this capability, we fine-tuned Qwen3-VL-4B-Instruct (QwenLM, 2025), a large-scale multimodal VLM that integrates a visual encoder, a multimodal projection (merger), and a language decoder. The fine-tuning process was designed to enhance the model's ability to perform context-aware visual reasoning and spatially grounded instruction generation based on architectural sketches. Specifically, we adopted a parameter-efficient supervised fine-tuning approach, in which the visual encoder and language model (LLM) were kept frozen while only the multimodal projection layer was updated. This strategy allowed the model to adapt its visual-textual alignment to the architectural sketch domain effectively, while mitigating the risks of overfitting and reducing computational cost.

Each training sample was a two-turn instruction–response conversation, simulating an interactive design process between a human designer and an intelligent assistant. The two-turn process consists of a detection phase, where the model identifies structural components, followed by a guidance phase, where the model generates renovation suggestions. As shown in Figure 2, in the first turn, the model receives an input sketch along with an instruction "Detect all windows and doors in the image". The model then performs structural understanding and spatial analysis, generating a textual interpretation that describes the building components and spatial zones, thereby forming a holistic understanding of the existing facade structure. In the second turn, the subsequent instruction builds upon the previous analysis, for example, "Update the layout based on the detected boxes". Conditioned on the prior context, the model indicates where new or modified elements should be introduced. Through this two-turn reasoning process, the model learns to associate structural understanding with renovation suggestion, enabling context-aware generation of coherent and spatially consistent textual guidance for facade modification.

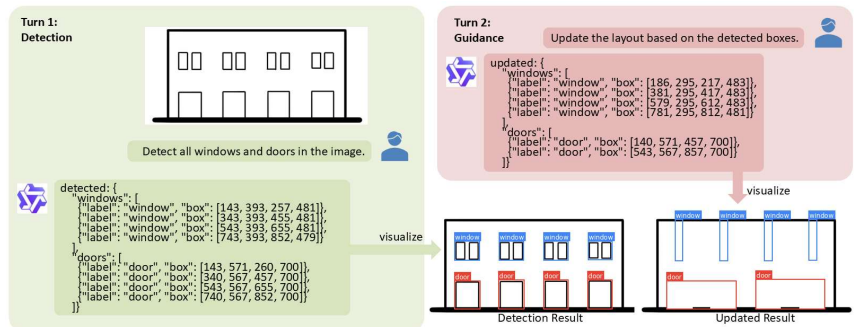

*Figure 2. Pipeline of the two-turn supervised fine-tuning process for the VLM*

## 3.2. COMPONENT GENERATION AND SKETCH ENHANCEMENT

Building upon the renovation guidance produced by the VLM, the second stage focuses on generating new structural components and enhancing the original sketch through a diffusion-based image generation process.

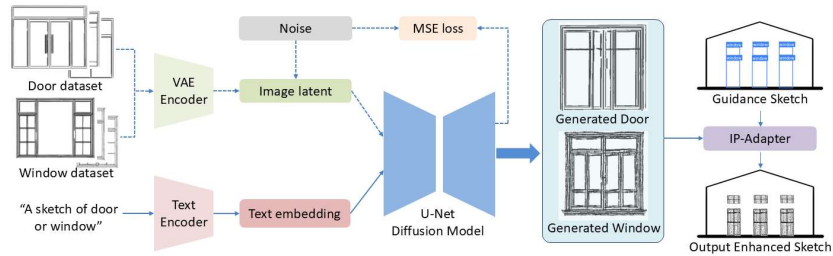

*Figure 3. Diffusion-based component generation and sketch enhancement pipeline*

As illustrated in Figure 3, a fine-tuned stable diffusion model (Rombach et al., 2022) serves as the generative backbone to synthesize architectural components such as windows and doors. Guided by the textual renovation descriptions from the previous stage, the diffusion model generates structurally aligned and stylistically consistent components, ensuring that newly created elements correspond precisely to the identified modification regions while maintaining the visual characteristics of the original sketch. Next, to integrate the generated components into the building outline, a pretrained IP-Adapter inpainting pipeline (Ye et al., 2023) is employed. The IP-Adapter extends the diffusion model with image-prompt conditioning, enabling localized editing and seamless blending between new and existing structures. The resulting enhanced sketch preserves geometric integrity and stylistic coherence, visually combining the existing facade with newly designed architectural features to reflect the intended renovation outcome.

## 3.3. PHOTOREALISTIC ARCHITECTURE IMAGE GENERATION

In the final stage, the enhanced renovation sketch, which combines the existing structural outline with newly generated design components, serves as the structural condition for photorealistic image synthesis. This stage aims to translate the conceptual renovation sketch into a high-fidelity architectural visualization, enabling realistic assessment of the proposed design.

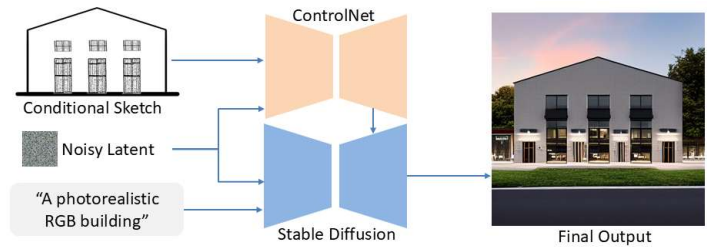

*Figure 4. Photorealistic image generation guided by sketch-based structural conditioning*

As shown in Figure 4, a pretrained stable diffusion model (Rombach et al., 2022) is adopted as the generative backbone, combined with a ControlNet module (Zhang et al., 2023) that enables spatial conditioning based on the input sketch. ControlNet constrains the diffusion process to follow the sketch structure, ensuring that the generated image preserves the original building geometry while achieving photorealistic rendering with appropriate materials, lighting, and textures. This structure-guided diffusion bridges the gap between conceptual sketches and realistic architectural visualization, supporting effective architectural review and presentation.

## 4. Experiments

### 4.1. TRAINING DATASET

#### 4.1.1. Facade Renovation Dataset

To fine-tune the VLM for renovation guidance, a specialized facade dataset was constructed focusing on industrial buildings in Tianjin, China, primarily built between the 1950s and 1980s. Building references were collected from publicly available imagery and renovation case archives, providing representative examples of industrial facade transformations. Based on these references, 100 paired facade samples (Figure 5) were manually created, each consisting of a "before-renovation" and an "after-renovation" image that correspond in geometry and transformation intent. Considering the variations in scale and roof morphology among heavy industrial factories, the dataset was categorized by building size and roof type. All selected cases represent steel-structure industrial buildings, which are generally in sound structural condition and offer high renovation potential. The buildings range from 10–20 meters in width and 5–10 meters in height, encompassing both pitched and flat roof configurations.

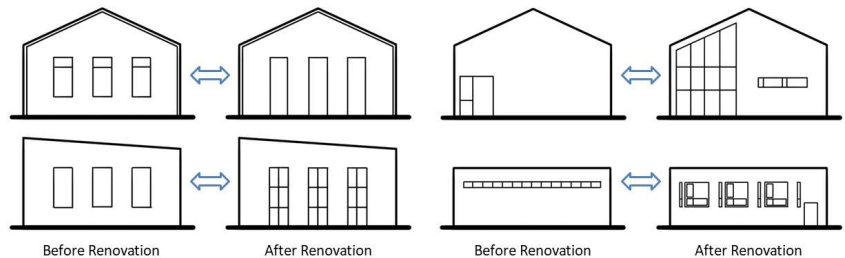

*Figure 5. Examples from the facade renovation dataset used for VLM fine-tuning*

#### 4.1.2. Component Dataset for Sketch Enhancement

To support the generation of architectural components during sketch enhancement, an auxiliary component dataset was also developed. As shown in Figure 6, this dataset comprises 107 door sketches and 164 window sketches, which serve as targeted training samples for the diffusion-based component generation model described in Section 3.2. Each sketch was manually drawn to capture representative geometries, and styles of facade elements commonly found in industrial renovation projects.

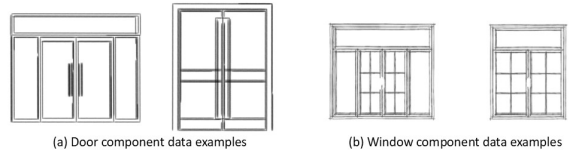

*Figure 6. Examples from the component dataset used for diffusion-based generation*

### 4.2. EXPERIMENTS

To evaluate the overall performance of the proposed three-stage generative framework, we conducted two types of experiments: reconstruction and generation. All experiments follow the same three-stage pipeline described in Section 3.

*4.2.1. Reconstruction Experiments*

In the reconstruction experiments, a subset of sketches from the training dataset was used as input to evaluate the framework's ability to reproduce design outcomes consistent with the training distribution. The VLM first generated textual renovation guidance, describing where and how facade components should be modified. Subsequently, the diffusion model synthesized corresponding architectural components (e.g., doors and windows) based on the VLM's suggestions, enhancing the input sketches. Finally, the ControlNet refined the enhanced sketches into photorealistic facade renderings. This process demonstrates the framework's ability to reproduce the geometric and stylistic consistency of the training data. Representative results are shown in Figure 7.

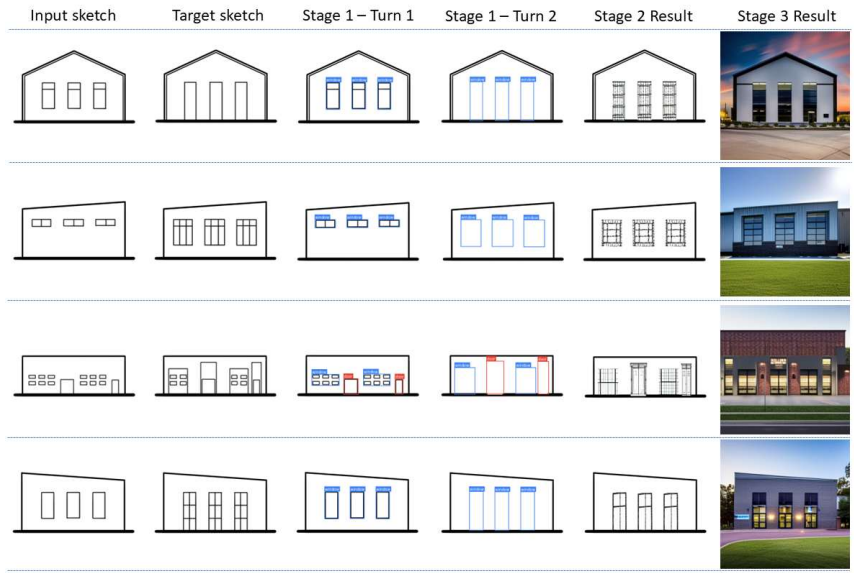

*Figure 7. Reconstruction results of the three-stage framework*

*4.2.2. Generation Experiments*

In the generation experiments, sketches that were not included in the training process were used as input to assess the framework's generalization ability. Following the same three-stage pipeline, the framework generated renovation suggestions, synthesized new facade components, and rendered photorealistic results. This experiment demonstrates that the proposed framework can generalize beyond the training distribution, producing design outcomes that are both semantically consistent and visually coherent for unseen building layouts. The generation results are presented in Figure 8.

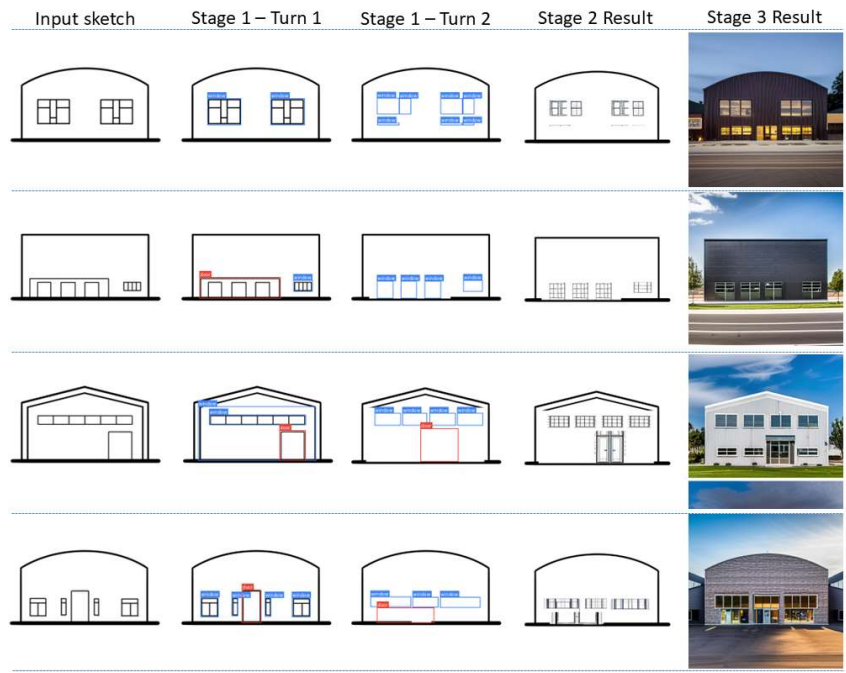

*Figure 8. Generation results of the three-stage framework on unseen sketches*

## 5. Results

### 5.1. RECONSTRUCTION RESULTS

As shown in Figure 7, the reconstruction results demonstrate that the proposed framework accurately reproduces renovation layouts that align with the geometric configuration of the original sketches. The generated outputs exhibit structural fidelity, with newly added components conforming to the existing facade proportions and spatial hierarchy. The results confirm that the fine-tuned VLM provides accurate semantic guidance, while the diffusion model and ControlNet collaboratively ensure visually realistic reconstruction from sketches to rendered images. These findings validate the framework's capability to maintain both semantic coherence and visual consistency across all stages of generation.

5.2. GENERATION RESULTS

As illustrated in Figure 8, the generation experiments demonstrate that the framework generalizes effectively to unseen architectural sketches. The model generates plausible and consistent renovation proposals, with component placement and scale aligned with the structural logic. This confirms that the combination of semantic reasoning (VLM), component synthesis (Diffusion), and visual refinement (ControlNet) enables handling new building configurations while producing convincing renovation outcomes.

5.3. REAL WORLD CASE STUDIES

To evaluate the applicability of the proposed three-stage generative framework, we collected photographs of existing industrial buildings and manually converted them into structural sketches that capture their primary geometric outlines and facade features. These sketches were then processed through the proposed generation pipeline, which produced renovation suggestions based on the model's semantic and spatial understanding. As shown in Figure 9, the framework demonstrates adaptability to real-world architectural conditions. The generated renovation guidance aligns closely with the existing facade geometry and architectural composition, maintaining spatial coherence while producing design suggestions that are both visually plausible and structurally consistent. These results highlight the framework's potential as a practical tool for facade renovation and architectural visualization in real-world design contexts.

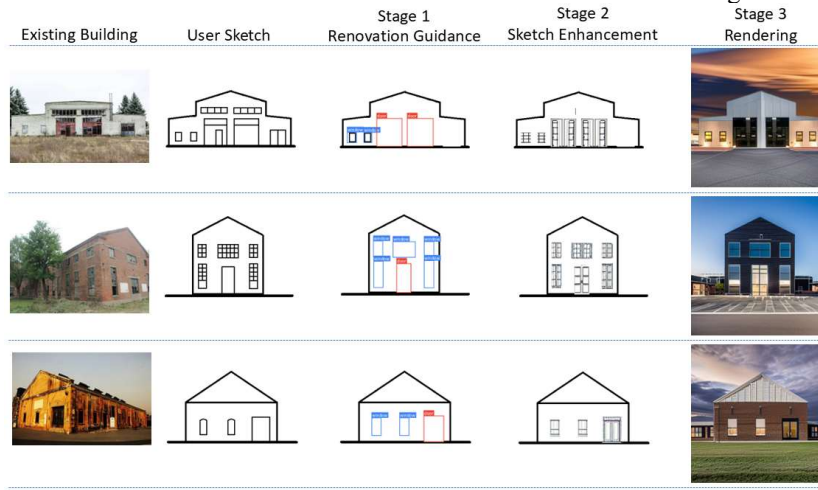

*Figure 9. Real-world case study results*

## 6. Conclusion

We proposed a novel sketch-based generative AI framework that transforms renovation concepts for existing industrial buildings into high-fidelity photorealistic visualizations, eliminating the need for time-consuming as-built modelling. Experimental results demonstrate strong potential in preserving structural integrity, enhancing facade detail, and improving visual realism in sketch-based renovation scenarios. The current framework focuses on simplified, front-facing building sketches and is designed to support early-stage visual exploration of renovation concepts. Future work will extend the training data and evaluation scope to more diverse architectural typologies, complex geometries, and viewpoints, and investigate quantitative evaluation protocols to better assess generalization performance. In addition, future research will explore the integration of structural and contextual constraints into the VLM-guided design process, so that the generated results can provide more reliable support for renovation decision-making. Through these extensions, we aim to advance the framework beyond the proof-of-concept stage toward a robust, deployable system that can support real-world architectural design and renovation workflows.

## Acknowledgements

This work was supported by JST SPRING, Japan Grant Number JPMJSP210, JST BOOST Program Japan Grant Number JPMJBY24D6, the National Natural Science Foundation of China, Grant Number 52508023, and the China Scholarship Council.

## Attribution

ChatGPT (OpenAI, 2025) was used to edit the manuscript and to improve flow.